\pdfoutput=1
\documentclass[sigconf]{acmart}

\usepackage{multirow}
\usepackage{balance}
 \usepackage{footnote}
\newcommand{\rmnum}[1]{\romannumeral #1}
\AtBeginDocument{%
  }

\copyrightyear{2023} 
\acmYear{2023} 
\setcopyright{acmlicensed}
\acmConference[MM '23] {Proceedings of the 31st ACM International Conference on Multimedia}{October 29--November 3, 2023}{Ottawa, ON, Canada.}
\acmBooktitle{Proceedings of the 31st ACM International Conference on Multimedia (MM '23), October 29--November 3, 2023, Ottawa, ON, Canada}
\acmPrice{15.00}
\acmISBN{979-8-4007-0108-5/23/10} 
\acmDOI{10.1145/3581783.3612490}

\begin{document}

\title{Seeing in Flowing: Adapting CLIP for Action Recognition \\with Motion Prompts Learning}

\author{Qiang Wang}
\affiliation{%
  \institution{Tencent YouTu Lab}
  \city{Shanghai}
  \country{China}}
\email{albertqwang@tencent.com}

\author{Junlong Du}
\affiliation{%
  \institution{Tencent YouTu Lab}
  \city{Shanghai}
  \country{China}}
\email{jeffdu@tencent.com}

\author{Ke Yan}
\authornote{Corresponding author}
\affiliation{%
  \institution{Tencent YouTu Lab}
  \city{Shanghai}
  \country{China}}
\email{kerwinyan@tencent.com}

\author{Shouhong Ding}
\affiliation{%
  \institution{Tencent YouTu Lab}
  \city{Shanghai}
  \country{China}}
\email{ericshding@tencent.com}

\begin{abstract}
  The Contrastive Language-Image Pre-training (CLIP) has recently shown remarkable generalization on ``zero-shot'' training and has applied to many downstream tasks. We explore the adaptation of CLIP to achieve a more efficient and generalized action recognition method. We propose that the key lies in explicitly modeling the motion cues flowing in video frames. To that end, we design a two-stream motion modeling block to capture motion and spatial information at the same time. And then, the obtained motion cues are utilized to drive a dynamic prompts learner to generate motion-aware prompts, which contain much semantic information concerning human actions. 
  In addition, we propose a multimodal communication block to achieve a collaborative learning and further improve the performance. 
  We conduct extensive experiments on HMDB-51, UCF-101, and Kinetics-400 datasets. Our method outperforms most existing state-of-the-art methods by a significant margin on ``few-shot'' and ``zero-shot'' training. We also achieve competitive performance on ``closed-set'' training with extremely few trainable parameters and additional computational costs.
\end{abstract}




\maketitle

\section{Introduction}
\label{sec:intro}
With the rapid increase of videos on the Internet, large-scale action recognition has become a critical problem that needs to be solved urgently. A powerful action recognition method should understand the semantic information in videos, even automatically describe the contents, \textit{e.g.,} human actions and complex events, and achieve an accuracy comparable to that of humans.

During the past few years, the developments of deep neural networks \cite{li2020tea,lin2019tsm,liu2021tam,qiu2017learning,tran2015learning,tran2018closer,xie2018rethinking} and current transformers \cite{arnab2021vivit,bertasius2021space,fan2021multiscale,girdhar2021anticipative,liu2022video,neimark2021video,yan2022multiview} have achieved promising progress in action recognition. However, most existing methods still suffer from two drawbacks regarding \textit{efficiency} and \textit{generalization}. 
On the one hand, action recognition is especially data-hungry for rare categories as for the difficulty of collection. Meanwhile, locating and annotating various human actions in original videos consumes much human effort. Therefore, the ``few-shot'' learning ability is essential for the efficient deployment of action recognition. On the other hand, most existing methods merely excel in dealing with the ``closed-set'' classification problem, in which all the categories are pre-defined and visible to the model. However, these methods are challenging to handle the unseen classes, limiting the practical applications, \textit{e.g.,} sports analysis \cite{selva2022video}, autonomous driving \cite{herath2017going}, and so on. Fortunately, recent research in image classification \cite{zhou2022learning,zhou2022conditional} has demonstrated that steering the large-scale Contrastive Language-Image Pre-training (CLIP) \cite{radford2021learning} to tackle classification tasks can significantly enhance the generalization of existing models. CLIP learns the joint representations from web-scale paired texts and images, then aligns the representations to a shared embedding space by simple noisy contrastive learning. As a result, the models equipped with CLIP show a remarkable ``zero-shot'' ability to recognize unseen categories in various image classification tasks \cite{zhou2022learning,zhou2022conditional,ju2021prompting}.

As for video domain, the natural idea is to train a video-language pre-trained model in the same way. However, constructing a web-scale video dataset takes up considerable storage resources compared to the image. In addition, affected by the irrelevant content in the videos, the textual descriptions and the videos on the web are permanently misaligned. Alternatively, another choice is to utilize a ``fine-tune'' manner, which turns the pre-trained parameters of CLIP into action recognition tasks. However, the ``fine-tune'' procedure unavoidably hurts the ``zero-shot'' generalization, leading to the degeneration of the CLIP. By contrast, a more economical and generalized way is to adapt the image representation generated by the image-language pre-trained model to video-level via extra modules. 
In this paper, we propose that the key lies in \textit{making CLIP see in flowing}, \textit{i.e.}, modeling the motion cues flowing in the frames to bridge the gap between still images and videos. To that end, we propose a two-stream Motion Modeling Block (MMB) to capture both the short- and long-term motion cues from representation differences between frames and the spatial features across all frames at the same time. As a result, our method yields a reconstructed video-level representation but maintains the generalization of CLIP via freezing the parameters of the image encoder. 

In addition, the second problem against the adaptation of CLIP is the gap between the category labels corresponding to human actions and the text documents used for training CLIP. To that end, the prompts engineering is introduced to form the category labels as ``fill-in-the-blank'' cloze tests \cite{petroni2019language}. For instance, compared to the original category label ``walk'', the text composed of the hand-crafted prompts ``\emph{human} \emph{action} \emph{of} [walk].'' is closer to the natural language description. It thus contains much semantic information and yields a more familiar textual input for the text encoder of CLIP. However, static prompts are inadequate as the lack of diversity. Hence, a prompts learning technique is further introduced to automatically generate dynamic prompts under optional conditions \cite{zhou2022conditional,ni2022expanding}. Recent research has already demonstrated the effectiveness of prompts learning. In this paper, we propose that the key lies in \textit{teaching prompts learner to describe actions}, that means the dynamic prompts should be generated under the guidance of motion cues concerning specific human action. Namely, the captured motion cues mentioned before is adopted as a signal and then fed into the prompts learner to yield Motion-Aware Prompts~(MAP). As a result, the aids of motion cues increase the semantic discriminativeness of the dynamic prompts regarding human actions, allowing our method to exhaustively explore semantic expression ability of CLIP.

Lastly, as mentioned before, we consider using extra modules to reconstruct the image-level representation to video-level with motion modeling. However, the exploration of motion modeling projects the original image representations to a new space, which potentially increases the difficulty to match the video and text representations and limits the performance. Here we propose a \textit{pre-matching} process via building a cross-modal communication between video and text representations. To that end, we propose a light-weight Multimodal Communication Block~(MCB) with two types of cross-modal attention, which aims to assign cross-modal prefixes for both the text and video representations and aid the matching of them. The experiments demonstrate that this collaborative learning further improves the performance.

We conduct comprehensive experiments on $3$ popular datasets, \textit{i.e.,} HMDB-51 \cite{kuehne2011hmdb}, UCF-101 \cite{soomro2012ucf101} and Kinetics-400 \cite{kay2017kinetics}. We adopt two training settings of ``few-shot'' and ``zero-shot'' to verify the \textit{efficiency} and the \textit{generalization} of our method, respectively. Then we compare the performance on par with the state-of-the-art methods under a ``closed-set'' setting. To summarize, the contributions of this paper are three-fold:
\begin{itemize}
  \item This paper introduces an explicit formulation of motion into the prompts learning of CLIP. The captured motion information yields a more generalized video-level representation on top of the frame-level features and steers a dynamic prompts learner to describe the human actions.
  \item We investigate a pre-matching process through a light-weight multimodal communication block in the adaptation of CLIP, which assigns cross-modal prefixes for both the text and the reconstructed video representations to guide the final matching process.
  \item Our method outperforms most state-of-the-art methods on ``few-shot'' and ``zero-shot'' training and achieves competitive ``closed-set'' Top-1 accuracy against most existing methods on three datasets with extremely few trainable parameters and extra computational costs.
\end{itemize}

\section{Related Work}
\noindent\textbf{Image-Language Models} The alignment of image and text is a traditional topic that has been studied last decade of years \cite{jia2021scaling,radford2021learning,zhang2020contrastive}. These methods usually focus on two aspects: \rmnum{1}) text and image representations engineering; \rmnum{2}) mapping text and image representations to a shared embedding space for distance measuring. Previous works usually utilize the hand-crafted feature descriptors \cite{elhoseiny2013write,socher2013zero} or deep neural networks \cite{frome2013devise,lei2015predicting} to generate the image embedding while using the pre-trained word vectors \cite{frome2013devise,socher2013zero} to obtain the text embedding. Then, extra constraint, formulated as metric learning \cite{frome2013devise}, multi-label classification \cite{gomez2017self}, or n-gram language learning \cite{li2017learning}, is adopted for cross-modality embedding alignment. A recent trend is to jointly train two encoders for text and image and utilize noisy contrastive learning to yield a generalized visual representation under the supervision of textual semantic information. A representative approach is CLIP \cite{radford2021learning}, which is trained using web-scale text and image pairs and showing remarkable ``zero-shot'' learning ability in image classification task \cite{zhou2022conditional,zhou2022learning}.

\begin{figure*}
    \centering
    \includegraphics[width=6.8in, keepaspectratio]{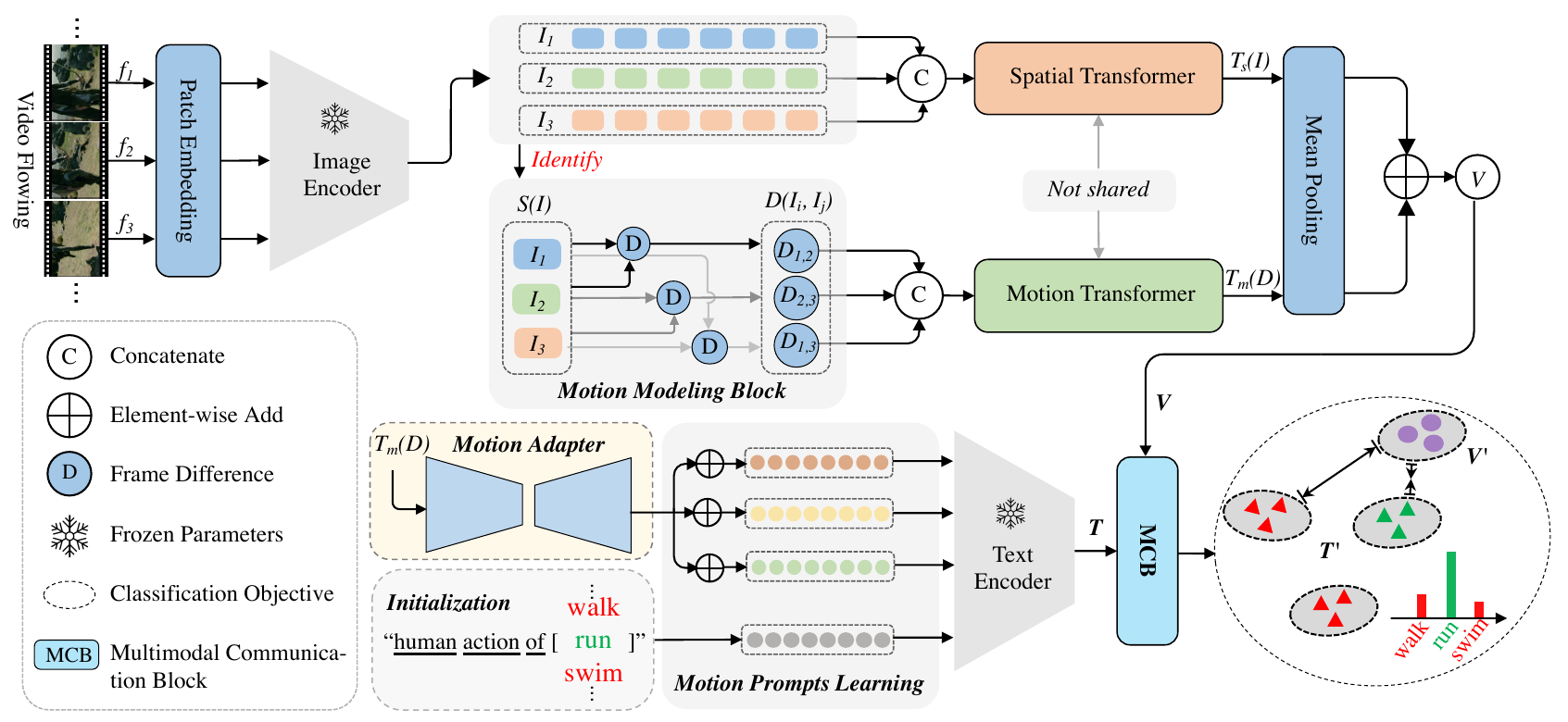}
    \caption{An overview of framework. We obtain the image representations of frames sampled from the video flow, and construct a video representation via modeling the motion cues. The motion information further steers the prompts learner to generate motion-aware prompts. Then, the multimodal communication block achieves a pre-matching process. The objective of our method is to maximize the similarity of cross-modal representations when the input video matches the category label.}
    \label{fig:framework}
\end{figure*}

\noindent\textbf{Prompts Learning} The idea of prompts learning is adopted from Natural Language Processing (NLP) domain. A cloze style template, \textit{a.k.a,} the prompts, is introduced to reconstruct the task as ``fill-in-the-blank'' cloze tests \cite{petroni2019language} and induce the pre-trained language model to generate appropriate answers. In practice, there are two mainstreams for prompts engineering. One is to design static prompts manually. For instance, GPT-3 \cite{brown2020language} and CLIP \cite{radford2021learning} benefit greatly from several hand-crafted prompts in ``zero-shot'' training. However, the design of hand-crafted prompts requires too much expert knowledge and the performance is always volatile. To that end, the second mainstream, called prompts learning, is proposed to generate dynamic prompts automatically. 
AutoPrompt \cite{shin2020autoprompt} proposes a gradient-based method to select the token leading to the most significant changes in gradients from a pre-defined vocabulary. By contrast, continuous prompt learning methods \cite{ju2021prompting,rao2022denseclip,zhang2022pointclip,zhou2022conditional} turns the static prompts into a set of learnable vectors that are optimized in an end-to-end manner. The prompts learning bridges the gap between pre-trained language models and downstream tasks significantly, sparking extensive exploration in the computer vision domain. CoCoOp \cite{zhou2022conditional} introduces the instance-level image representation into the procedure of prompts generation, benefiting the ``few-shot'' and ``zero-shot'' training. Leveraging a context-aware prompting method, DenseCLIP \cite{rao2022denseclip} transfers the pre-trained CLIP model into dense prediction tasks, such as semantic segmentation and object detection, while X-CLIP \cite{ni2022expanding} designs the video-specific prompts to expand the image-language pre-trained model into action recognition tasks.

\noindent\textbf{Action Recognition} As a fundamental task in the vision domain, action recognition aims to identify human actions in videos. Deep learning-based methods have undergone rapid developments, which are divided into two categories according to architecture. CNN-like methods usually utilize a two-stream network  \cite{li2020tea,lin2019tsm,liu2021tam,qiu2017learning,tran2015learning,tran2018closer,xie2018rethinking} to model spatial features from static images and temporal information from optical flow, respectively. In addition, ViT-like methods \cite{arnab2021vivit,bertasius2021space,fan2021multiscale,girdhar2021anticipative,liu2022video,neimark2021video,yan2022multiview}, usually consisting of a frame-level spatial transformer and a temporal fusion module, achieve more promising performance with the aid of abundant labeled training data. Recently, image-language pre-trained models are also applied to action recognition tasks. For instance, ActionCLIP \cite{wang2021actionclip} proposes a new paradigm of \textit{pre-train, prompt and finetune} for action recognition, while Ju~\textit{et al.} \cite{ju2021prompting} expands CLIP into several video understanding tasks via prompt learning. Benefiting from the powerful generalization of CLIP, these methods perform excellently on ``few-shot'' and ``zero-shot'' training.

\section{Methods}
In this section, we first briefly overview the framework of our method and then introduce our three key components, \textit{i.e.,} the video encoder, the text encoder, and the multimodal communication block in detail. 

\subsection{Overview}
We represent our framework in Fig.~\ref{fig:framework}. Our proposed method is built upon the approach of CLIP \cite{radford2021learning}. A ViT-like CLIP model takes two parallel transformers \cite{vaswani2017attention} to generate text and image representations, respectively. In this paper, we expand the original encoders of CLIP with light-weight extensions to transfer the pre-trained image-language model into action recognition task. The two encoders of CLIP are 
converted to a novel \textit{video encoder} and a \textit{text encoder}, which are introduced in Sec.~\ref{sec3.2} and Sec.~\ref{sec3.3} in detail, respectively. Here, we represent the video encoder and the text encoder of our method as $\mathcal{F}_{V}$ and $\mathcal{F}_{T}$ for brevity. Given a video clip \textit{V} consisting of $T$ static frames and a set of $K$ corresponding category labels $\mathcal{C}$, the text encoder generates the textual representation $\mathcal{T}_{i}$ for $i$-th category, while the video encoder obtains a video-level representation $\mathcal{V}$, namely that:
\begin{equation}
    \label{eq1}
    \begin{aligned}
        \mathcal{T} = \{\mathcal{T}_{i}~|~& 1~\le~i~\le~K\}, \mathcal{T}_{i} = \mathcal{F}_{T}(\mathcal{C}_{i}),\\
        & \mathcal{V} = \mathcal{F}_{V}(V).
    \end{aligned}
\end{equation}

And then, two representations $\mathcal{T}$ and $\mathcal{V}$ are fed into our proposed Multimodal Communication Block (MCB) to integrate multimodal representations and achieve collaborative learning during training. The computation is formally represented as follows:
\begin{equation}
    \label{eq2}
    \begin{aligned}
        \mathcal{T}',~\mathcal{V}' = \mathrm{MCB}(\mathcal{T},~\mathcal{V}),
    \end{aligned}
\end{equation}
we show more details of MCB in Sec.~\ref{sec3.4}. At last, the probability that the video representation $\mathcal{V}'$ matches the text representation of $i$-th category $\mathcal{T}_{i}'$ is computed as:
\begin{equation}
    \label{eq3}
    \begin{aligned}
        p(\mathcal{T}'_{i},~\mathcal{V}') = \frac{\mathrm{exp}(\langle~\mathcal{T}'_{i}, \mathcal{V}'\rangle~/~\tau)}{\sum_{j=1}^{K}\mathrm{exp}(\langle~\mathcal{T}'_{j},~\mathcal{V}'\rangle~/~\tau)},
    \end{aligned}
\end{equation}
where $\tau$ is a temperature hyper-parameter and $\langle\cdot,\cdot\rangle$ denotes the cosine similarity. We utilize a NCE loss as the objective function to maximize the probability $p(\mathcal{T}'_{i},~\mathcal{V}')$ when $V$ matches $\mathcal{C}_{i}$.
% \begin{equation}
%     \label{eq3-2}
%     \begin{aligned}
%         \mathcal{L}=\sum_{i}^{K}\mathrm{log}p_{i}.
%     \end{aligned}
% \end{equation}

\subsection{Video Encoder}
\label{sec3.2}
Our proposed video encoder mainly consists of two elements: \rmnum{1}) the frozen image encoder of CLIP aiming to extract spatial features of video frames. \rmnum{2}) a novel Motion Modeling Block (MMB) to integrate both the motion and spatial features and yield a more abundant video-level representation, which is more appropriate for the action recognition task.

\noindent\textbf{Frame Representations} Given a video clip $V\in\mathbb{R}^{T\times H\times W\times 3}$ consisting of $T$ frames with a spatial resolution of $H\times W$, the $t$-th frame is first divided into $N$ patches with $P\times P$ pixels, and then the patches are projected to a collection of patch embeddings $\mathcal{E}_{t} = \{ e_{i}\in \mathbb{R}^{M}~|~1\le i \le N\}$ by a linear projection, where $N=H\times W/P^{2}$ and $M$ is the dimension of patch embeddings. Then a class embedding $e_{c}\in \mathbb{R}^{M}$ is appended to the head of $\mathcal{E}_{t}$. After that, the patch embeddings of $T$ frames are fed into the frozen image encoder of CLIP to generate frame-level visual representations $\mathcal{I} = \{\mathcal{I}_{t}\in\mathbb{R}^{D}~|~1\le t \le T\}$, where $D$ is the dimension of visual representations.

\noindent\textbf{Motion Modeling Block} The representations extracted from static frames are indeed not sufficient to perceive the movements flowing in videos. In other words, CLIP needs motion-aware guidelines. To that end, we propose a motion modeling block consisting of two streams. 
The \textit{motion stream} aims to capture motion cues from the differences of representations between video frames, while the \textit{spatial stream} achieves an integration crossing through the spatial features of all frames. Lastly, the motion cues and the integrated spatial features are merged to generate a video-level representation. 

Formally, we first define $\mathcal{S(I)}$ to select two different frame representations from $\mathcal{I}$ with a hyper-parameter $\textit{s}~(1 \le s \le T - 1)$, which denotes the maximum temporal interval between selected frame representations. It means that the motion modeling block considers both the short- and long-term temporal information. 
Then we compute a set of representation differences $\mathcal{D}\in\mathbb{R}^{D}$ as follows:
\begin{equation}
    \label{eq4}
    \begin{aligned}
        \mathcal{D} = \{\mathcal{D}_{k}=\mathcal{I}_{j} - \mathcal{I}_{i}~|i, j \in \mathcal{S(I)},~i~<~j\},
    \end{aligned}
\end{equation}
where $D$ is the representation dimension. Then we utilize two transformers, \textit{i.e.,} $\mathcal{T}_{m}$ and $\mathcal{T}_{s}$ to obtain the motion representation $\mathcal{M}$ and the cross-frame spatial features $\mathcal{S}$, respectively. Namely that:
\begin{equation}
    \label{eq5}
    \begin{aligned}
        \mathcal{M} &= \mathcal{T}_{m}(\mathcal{D})\in\mathbb{R}^{L\times D},\\
        \mathcal{S} &= \mathcal{T}_{s}(\mathcal{I})\in\mathbb{R}^{T\times D},
    \end{aligned}
\end{equation}
where $L=\sum_{i=1}^{s}T-i$ is the length of $\mathcal{D}$. The two transformers are constructed by the standard architecture in \cite{vaswani2017attention} including a Multi-Head Self-Attention (MHSA) and a Feed-Forward Network (FFN). At last, $\mathcal{M}$ and $\mathcal{S}$ are averaged via AvgPool operations and then aggregated to obtain the final video-level representation $\mathcal{V}\in\mathbb{R}^{D}$ as follows:
\begin{equation}
    \label{eq6}
    \begin{aligned}
        \mathcal{V} = \mathrm{AvgPool}(\mathcal{M}) + \mathrm{AvgPool}(\mathcal{S}).
    \end{aligned}
\end{equation}

% \begin{figure}
%     \centering
%     \includegraphics[width=3.2in, keepaspectratio]{text_encoder.pdf}
%     \caption{An overview of text encoder. We utilize a motion-aware prompts learning method to generate prompts that contain motion information, which is introduced in Sec.~\ref{sec3.3} in detail.
%     }
%     \label{fig:text_encoder}
% \end{figure}

\subsection{Text Encoder}
\label{sec3.3}
The text encoder $\mathcal{F}_{T}$ aims to generate diverse text representations for category labels leveraging motion-aware prompts learning. The video representation containing motion cues steers the optimization of the prompts learner, leading to specific motion-aware prompts for textual input.

\noindent\textbf{Motion-Aware Prompts} 
Formally, the prompts learner defines a set of learnable vectors $\{\mathcal{P}_{i}\in\mathbb{R}^{D}~|~1\le i\le H\}$, where $H$ is a hyper-parameters that denotes the length of tokens need to predict. The specific prompts $\mathcal{P}$ for category labels is constructed as follows:
\begin{equation}
    \label{eq7}
    \begin{aligned}
        \mathcal{P} = [\texttt{SOS}][\mathcal{P}_{1}][\mathcal{P}_{2}]\dots[\mathcal{P}_{H}][\mathrm{CLASS}][.][\texttt{EOS}], 
    \end{aligned}
\end{equation}
where the token sequence is capped at a fixed length of $77$ by two border tokens $[\texttt{SOS}]$ and $[\texttt{EOS}]$. The tokens of $\mathcal{P}$ are first converted to a vector of numeric IDs by a $\texttt{Tokenizer}$ \cite{sennrich2015neural} according to a look-up table. And then the vector is encoded to a collection of token embeddings $\mathcal{E}_{\mathcal{P}}\in\mathbb{R}^{77\times D}$. The prompts vectors $\mathcal{P}_{i}$ are first initialized by hand-crafted prompts, \textit{e.g.,} ``\emph{human} \emph{action} \emph{of} [CLASS].'' when $H$ equals $3$. Then the prompts vectors are optimized end-to-end under the guidelines of motion cues. As shown in Fig.~\ref{fig:framework}, we utilize a \textit{motion adapter} that convert the information obtained by MMB to motion-aware guidance. The motion cues $\mathcal{M}$ are first mapped by a linear projection and a non-linear activation to squeeze the dimension of visual features, and then another linear projection is followed to recover the feature to the same dimension with prompts vectors. At last, the projected motion information is aggregated to $\mathcal{P}_{i}$ by element summation. As a result, we obtain the updated token embeddings under the guidance of motion information.

\begin{figure}
    \centering
    \includegraphics[width=3.1in, keepaspectratio]{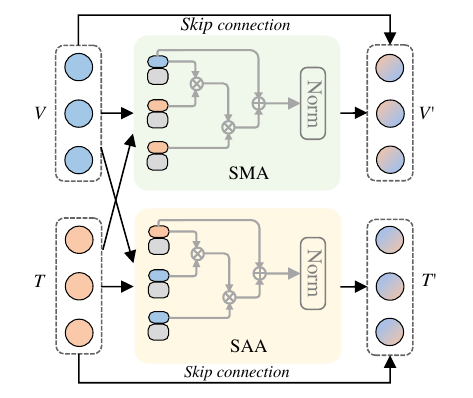}
    \caption{An illustration of multimodal communication block and the pre-matching process. MCB contains two types of cross-modal attention, \textit{i.e.,} the Semantic Matching Attention~(SMA) and the Semantic Allocating Attention~(SAA) to enhance the semantic perception via a collaborative learning during training. More details are shown in Sec.~\ref{sec3.4}.}
    \label{fig:mcb}
\end{figure}

\begin{figure*}
    \centering
    \includegraphics[width=6.8in, keepaspectratio]{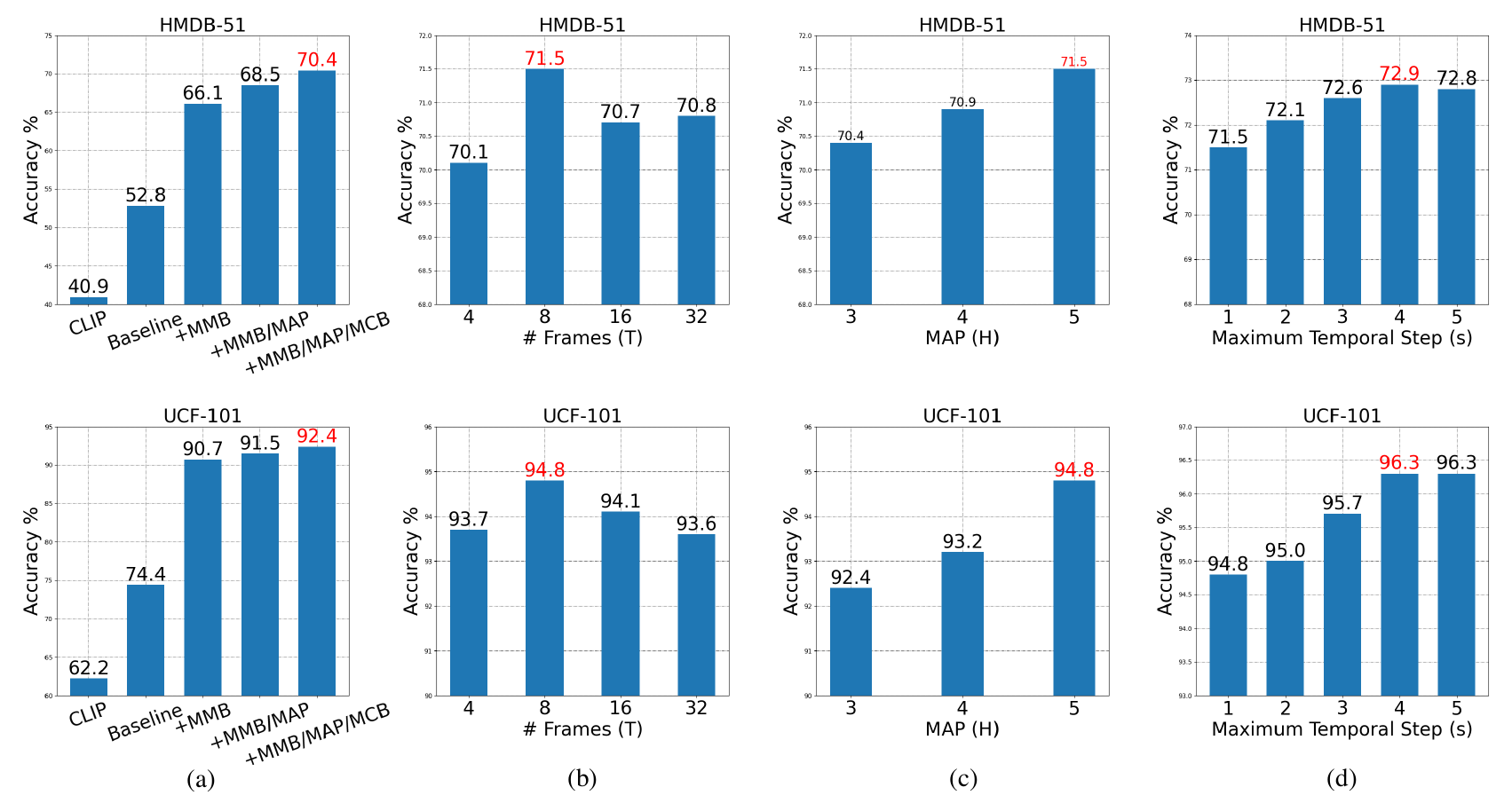}
    \caption{Ablation studies on HMDB-51 and UCF-101 datasets under ``close-set'' training. (a) The effects of MMB, MAP and MCB; (b) The number of sampled frames; (c) The length of learnable vectors in MAP. (d) The maximum temporal step \textit{s}.}
    \label{fig:ablation}
\end{figure*}

\noindent\textbf{Textual Representation} 
Finally, we feed the token embeddings into the frozen text encoder of CLIP and obtain a specific textual representation for each category label $\mathcal{T}=\{\mathcal{T}_{i}\in\mathbb{R}^{D}~|~1\le i\le K\}$, where $K$ is the number of categories. It needs to mention that the learnable prompts vectors are shared by all category labels, therefore, the difference in the textual representations is only related to the text of category labels. However, the motion representation provides extra information which aids the prompts learner to generate motion descriptions, increasing the semantic discriminativeness of text representations.

\subsection{Multimodal Communication Block}
\label{sec3.4}

As mentioned before, the additional projections within motion modeling block potentially increases the difficulty to match the video and text representations and limits our performance. Thus we propose to build a \textit{pre-matching} process via a light-weight Multimodal Communication Block~(MCB). 

As shown in Fig.~\ref{fig:mcb}, MCB contains two parallel attentions, which are named Semantic Matching Attention~(SMA) and Semantic Allocating Attention~(SAA), respectively. These two attentions achieve cross-modal information interaction through collaborative learning during training. Concretely, semantic matching attention tasks the text representation $\mathcal{T}$ as query while the video representation $\mathcal{V}$ as key and value, thus the explicit semantic information contained in natural languages attempts to match the corresponding visual details within videos. By contrast, the semantic allocating attention adopts $\mathcal{V}$ as a query while $\mathcal{T}$ as the key and value for allocating textual semantics to the visual representation. These two attentions generate cross-modal prefixes, which are injected into the original $\mathcal{T}$ and $\mathcal{V}$ to aid the final matching process. We construct the two attentions by the standard cross attention with residual connection.

\section{Experiments}
In this section, we conduct experiments on three popular datasets, \text{i.e.}, HMDB-51\cite{kuehne2011hmdb}, UCF-101\cite{soomro2012ucf101} and Kinetics-400\cite{kay2017kinetics}. We introduce the experimental details at first. And then we conduct ablation studies to verify the effectiveness of three key designs. The extensive experiments demonstrate the efficiency and the generalization of our method on ``few-shot'' and ``zero-shot'' training, respectively. Our best models on three datasets also achieve competitive Top-1 accuracy on par with most existing state-of-the-art methods.

\subsection{Experimental Protocols}
\label{sec_4_1}
\noindent\textbf{Datasets and Evaluation.} \textbf{HMDB-51} \cite{kuehne2011hmdb} dataset contains about $7,000$ videos of $51$ categories. We adopt three standard splits that $3,570$ and $1,530$ videos are used for training and testing, respectively. \textbf{UCF-101} \cite{soomro2012ucf101} dataset consists of around $13,000$ videos with regard to $101$ actions. Three standard splits that $9,537$ videos for training and $3,783$ videos for testing are adopted. \textbf{Kinetics-400}~(K-400) \cite{kay2017kinetics} contains more than $240,000$ videos collected from YouTube. Here, we use around $230,000$ videos for training and about $19,000$ videos for testing. Following the setting of CLIP, the resolution of frames is set to $224\times224$ for all datasets. 
We evaluate all of our models on $4\times3$ views, \textit{i.e.,} randomly sample $4$ clips in a video and crop frames to $224\times224$ pixels for $3$ times, then the average Top-1 and Top-5 accuracy of all views are reported.

\begin{table}
\setlength{\tabcolsep}{2mm}{
\begin{tabular}{cccccc}
\toprule
\multirow{2}{*}{Methods} & \multirow{2}{*}{Prompts} & \multicolumn{2}{c}{HMDB-51} & \multicolumn{2}{c}{UCF-101} \\ \cline{3-6} 
                         &                          & Top-1        & Top-5        & Top-1        & Top-5        \\ \hline
\multirow{3}{*}{CLIP}    & hand-crafted A           & 40.9         & 70.1         & 62.2         & 86.0           \\
                         & hand-crafted B           & 40.6         & 71.2         & 63.5         & 87.0         \\
                         & hand-crafted C           & 41.7         & 70.4         & 63.1         & 86.2         \\ \hline
\multirow{6}{*}{Ours}    & \multirow{2}{*}{MAP(H=3)}& 68.5         & 91.9         & 91.5         & 99.2         \\
                         &                          & +27.6        & +21.8        & +29.3        & +13.2        \\
                         & \multirow{2}{*}{MAP(H=4)}& 66.0         & 92.2         & 92.3         & 99.0         \\
                         &                          & +25.4        & +21.0        & +28.8        & +12.0        \\
                         & \multirow{2}{*}{MAP(H=5)}& 66.9         & 92.3         & 92.9         & 99.3         \\ 
                         &                          & +25.2        & +21.9        & +29.8        & +13.1        \\                         
                         \bottomrule
\end{tabular}
}
\caption{Motion-aware prompts vs. Hand-crafted prompts. We design three hand-crafted prompts for CLIP. Specifically, hand-crafted A denotes ``human action of [CLASS].'', hand-crafted B denotes ``a human action of [CLASS].'' and hand-crafted C is ``a common human action of [CLASS].''.}
\label{map_vs_clip}
\end{table}

\begin{table}
\setlength{\tabcolsep}{0.4mm}{
\begin{tabular}{ccccc}
\toprule
\multirow{2}{*}{Initialization of MAP(H=5)} & \multicolumn{2}{c}{HMDB-51} & \multicolumn{2}{c}{UCF-101} \\ \cline{2-5} 
 & Top-1 & Top-5 & Top-1 & Top-5 \\ \hline
 a common human action of {[}CLASS{]}.   & 71.5  & 93.2  & 93.9  & 99.3  \\
 the common human action of {[}CLASS{]}. & 71.4  & 93.1  & 93.7  & 99.4  \\
 a popular human action of {[}CLASS{]}.  & 71.5  & 93.3  & 93.8  & 99.4  \\ \bottomrule
\end{tabular}
}
\caption{Comparison of different initialization for MAP(H=5) on HMDB-51 and UCF-101 datasets. The performance is barely affected by the content of prompts for initialization. }
\label{init_map}
\end{table}

\begin{table}[]
\begin{tabular}{ccccc}
\toprule
\multirow{2}{*}{Methods} & \multirow{2}{*}{Params(MB)} & \multicolumn{3}{c}{GFLOPs}       \\ \cline{3-5} 
                         &                             & HMDB-51 & UCF-101 & Kinetics-400 \\ \hline
CLIP                     & 139.7                       & 459.470  & 555.366 & 1135.759     \\
Baseline                 & 141.8                       & 459.490  & 556.383 & 1135.776     \\ \hline
\multirow{2}{*}{Ours}    & 143.9                       & 459.499  & 556.397 & 1135.793     \\
                         & +4.2                        & +0.0290  & +0.031  & +0.034       \\
\bottomrule
\end{tabular}
\caption{Analysis of our trainable parameters and additional computational costs.}
\label{param_flops}
\end{table}

\noindent\textbf{Settings and Baseline.} In all experiments, we adopt the image and text encoders of CLIP-B/16~\cite{radford2021learning} and freeze the pre-trained parameters of two encoders during training. 
We set the dimension of representations $D$ and the dimension of path embeddings $M$ to $512$ and $768$, respectively. The temperature parameter $\tau$ in Eq.~\ref{eq3} is set to $0.07$. To analyze the key elements of our method, we design a slim ``Baseline'' model for ablation studies. The differences compared to ours are three-fold: \rmnum{1}) the ``Baseline'' model is built upon CLIP-B/16 with only one single Transformer to mimic the \textit{spatial stream} in our video encoder, while the \textit{motion stream} is removed. \rmnum{2}) the ``Baseline'' only receives hand-crafted prompts and generates static text representations. \rmnum{3}) MCB is not applied in ``Baseline''.

\subsection{Ablation Study}
Firstly, we conduct an ablation study to analyze our three key designs, \textit{i.e.,} the Motion Modeling Block (MMB), the Motion-Aware Prompts learning (MAP), and the Multimodal Communication Block (MCB), the number of sampled video frames $T$, the length of learnable vectors $H$ in Eq.~\ref{eq7} and the maximum temporal step \textit{s} under ``closed-set'' training. We train all models on $8$ Tesla-V100 cards with a batchsize of $128$ and an initial learning rate of $0.0025$ for both of HMDB-51 and UCF-101. A standard SGD optimizer and a cosine learning rate scheduler are utilized for optimization. The results are shown in Fig.~\ref{fig:ablation}.

\noindent\textbf{Key designs} In Fig.~\ref{fig:ablation}~(a), we evaluate the origin CLIP model with hand-crafted prompts and obtain a video representation via Mean Pooling operation across $8$ frames. 
Compared to ``Baseline'', our model merely equipped with motion modeling block improves Top-1 accuracy significantly, \textit{i.e.,} $+13.3\%$ and $+16.0\%$ improvements on two datasets. With the aid of the motion cues, the motion-aware prompts learning strategy with $H=3$ further improves $+2.4\%$ and $+0.8\%$ on HMDB-51 and UCF-101, respectively. In summary, motion cues matter. The results demonstrate a significant gap between the image and video domains and the importance of motion modeling. In addition, the multimodal communication block also shows positive effects, in which improves the Top-1 accuracy by $+1.9\%$ and $+0.9\%$ on HMDB-51 and UCF-101, verifying the effectiveness of the our pre-matching process.

\begin{table}[]
\setlength{\tabcolsep}{4mm}{
    \begin{tabular}{ccccc}
    % \hline
    \toprule
    Methods               & MMB & MAP & MCB & Top-1 \\ \hline
    Ju~\textit{et al.}~\cite{ju2021prompting} & - & - & - & 58.5  \\ \hline
    Baseline              &     &     &     & 54.8  \\
    \multirow{4}{*}{Ours} &     &     & $\checkmark$   & 58.4  \\
                          & $\checkmark$   &     &     & 57.9  \\
                          & $\checkmark$   & $\checkmark$   &     & 60.3  \\
                          &   $\checkmark$ &   $\checkmark$ &   $\checkmark$ &   \textbf{61.5}  \\ % \hline
    \bottomrule
    \end{tabular}
}
\caption{The results of $5$-shot training on Kinetics-400 dataset.}
\label{few_shot_k400}
\end{table}

\begin{table*}
\setlength{\tabcolsep}{3.6mm}{
    \begin{tabular}{cccccccccccc}
    % \hline
    \toprule
    \multirow{2}{*}{Methods} & \multirow{2}{*}{MMB} & \multirow{2}{*}{MAP} & \multirow{2}{*}{MCB} & \multicolumn{4}{c}{HMDB-51} & \multicolumn{4}{c}{UCF-101} \\ \cline{5-12} 
                          & - & - & - & \textit{k}=2  & \textit{k}=4  & \textit{k}=8  & \textit{k}=16 & \textit{k}=2  & \textit{k}=4  & \textit{k}=8  & \textit{k}=16 \\ \hline
    TSM~\cite{lin2019tsm} & - & - & - & 17.5 & 20.9 & 18.4 & 31.0 & 25.3 & 47.0 & 64.4 & 61.0 \\
    TimeSformer~\cite{bertasius2021space} & - & - & - & 19.6 & 40.6 & 49.4 & 55.4 & 48.5 & 75.6 & 83.7 & 89.4 \\
    Swin-B~\cite{liu2022video} & - & - & - & 20.9 & 41.3 & 47.9 & 56.1 & 53.3 & 74.1 & 85.8 & 88.7 \\
    X-CLIP~\cite{ni2022expanding} & - & - & - & 53.0 & 57.3 & 62.8 & 64.0 & 76.4 & 83.4 & 88.3 & 91.4 \\ \hline
    Baseline              &   &   &   & 35.7 & 39.0 & 42.1 & 55.8 & 64.0 & 73.3 & 78.2 & 80.6 \\
    \multirow{4}{*}{Ours} &   &   & $\checkmark$ & 49.4 & 53.8 & 57.4 & 59.4 & 72.3 & 77.0 & 81.1 & 84.4 \\
                          & $\checkmark$ &   &   & 53.1 & 55.4 & 58.4 & 60.4 & 77.0 & 77.7 & 81.5 & 83.9 \\
                          & $\checkmark$ & $\checkmark$ &   & 54.2 & 57.4 & 62.1 & 62.9 & 78.4 & 81.0 & 85.6 & 87.8 \\
                        %   \rowcolor{blue!8}
                          &   $\checkmark$ &   $\checkmark$ &   $\checkmark$ &   \textbf{55.3} &   \textbf{58.7} &   \textbf{64.0} &   \textbf{64.6} &   \textbf{82.4} &   \textbf{85.8} &   \textbf{89.1} &   \textbf{91.6} \\ %\hline
    \bottomrule
    \end{tabular}
}
\caption{Few-Shot training on HMDB-51 and UCF-101 datasets. We adopt a standard \textit{k}-shot setting and set the \textit{k} to $2,4,8,16$, respectively. The averaged Top-1 accuracy of $10$ runs is reported. }
\label{few_shot_hmdb_ucf}
\end{table*}

\noindent\textbf{Number of video frames} We discuss the effects of the number of video frames $T$ in Fig.~\ref{fig:ablation}~(b). We find that the increase of $T$ diminishes the performance gain on both HMDB-51 and UCF-101. We compare the results of $T\in\{4,~8,~16,~32\}$, and the best performance is achieved when $T$ equals $8$. We conjectured that the frozen CLIP image encoder introduces an \textit{inner variance} crossing video frames, which external parameters can not eliminate. A larger number of frames contain much irrelevant noise, leading to a more significant variance that hurts the performance. Therefore, we set $T$ to $8$ in the following experiments. 

\noindent\textbf{Length of learnable vectors} As shown in Fig.~\ref{fig:ablation}~(c), we set $H\in\{3,~4,~5\}$ and compare the performance under the same setting. The results show that the increase of $H$ can further boost performance. For instance, increasing $H$ from $3$ to $5$, our models equipped with MMB, MAP and MCB gain $1.1\%$ and $2.4\%$ on HMDB-51 and UCF-101, respectively. That is to say, a larger length of the learnable vectors within MAP expands the semantic capacity of dynamic prompts, leading to more explicit descriptions regarding human actions. In the following experiments, we set $H=5$ to achieve a trade-off between performance and training budgets. 
% In addition, our best models exceed Ju~\textit{et al.}~\cite{ju2021prompting} by $+5.1\%$ of Top-1 and $+1.1\%$ of Top-5 on HMDB-51, while $+1.2\%$ of Top-1 and $+0.3\%$ of Top-5 on UCF-101.

\begin{table}[]
\setlength{\tabcolsep}{5.8mm}{
    \begin{tabular}{ccc}
    % \hline
    \toprule
    Methods    & HMDB-51  & UCF-101  \\ \hline
    MTE~\cite{xu2016multi} & 19.7~$\pm$~1.6 & 15.8~$\pm$~1.3 \\
    ASR~\cite{wang2017alternative} & 21.8~$\pm$~0.9 & 24.4~$\pm$~1.0 \\
    ZSECOC~\cite{qin2017zero} & 22.6~$\pm$~1.2 & 15.1~$\pm$~1.7 \\
    UR~\cite{zhu2018towards} & 24.4~$\pm$~1.6 & 17.5~$\pm$~1.6 \\
    TS-GCN~\cite{gao2019know} & 23.2~$\pm$~3.0 & 34.2~$\pm$~3.1 \\
    E2E~\cite{brattoli2020rethinking} & 32.7~$\pm$~0.0 & 48.0~$\pm$~0.0 \\
    ER-ZSRA~\cite{chen2021elaborative} & 35.3~$\pm$~4.6 & 51.8~$\pm$~2.9 \\
    ActionCLIP~\cite{wang2021actionclip} & 40.8~$\pm$~5.4 & 58.3~$\pm$~3.4 \\
    X-CLIP~\cite{ni2022expanding} & 44.6~$\pm$~5.2 & 72.0~$\pm$~2.3 \\ 
    Vita-CLIP~\cite{wasim2023vita} & 48.6~$\pm$~0.6 & 75.0~$\pm$~0.6 \\
    \hline
    \textbf{Ours} & \textbf{50.1~$\pm$~5.4} &   \textbf{76.4~$\pm$~2.5}  \\ %\hline
    \bottomrule
    \end{tabular}
}
\caption{Zero-Shot training on K-400 dataset. The pre-trained model is adapted to HMDB-51 and UCF-101 datasets. The average Top-1 accuracy and the standard deviation over three standard splits are reported. }
\label{zero_shot_k400}
\end{table}

\begin{table}[]
\setlength{\tabcolsep}{5mm}{
\begin{tabular}{ccc}
% \hline
\toprule
Zero-Shot Training & Baseline & Ours \\\hline
HMDB-51~→~UCF-101 & 33.8~$\pm$~3.5 & \textbf{62.7~$\pm$~2.8} \\
UCF-101~→~HMDB-51 & 16.6~$\pm$~1.9 & \textbf{38.4~$\pm$~2.5} \\
% \hline
\bottomrule
\end{tabular}
}
\caption{Zero-shot training on HMDB-51 and UCF-101 datasets. We apply our model trained on HMDB-51 to UCF-101 without any turning and vice versa. We report the average Top-1 and standard deviation on $3$ standard splits.}
\label{zero_shot_hmdb_ucf}
\end{table}

\noindent\textbf{The maximum temporal step} The setting of the maximum temporal step \textit{s} balances the effects of shot- and long-term temporal information. Here we set $s\in\{1,~2,~3,~4,~5\}$. The results in Fig.~\ref{fig:ablation}~(d) show that: \rmnum{1} compared to $s=1$ in which merely considers adjacent frames, the performance of $s > 1$ is better, which means that the long-term motion cues are also important. \rmnum{2} The performance hardly improves when $s$ exceeds $4$. The reasons are two-fold. A large $s$ suppresses the effect of short-term motion cues, and leads to more information redundancy. Thus we set the $s$ to $4$ in the following experiments.

\noindent\textbf{Motion-aware prompts vs. Hand-crafted prompts} We compare the effects of motion-aware and hand-crafted prompts. We design three hand-crafted prompts in the length of ${3,~4,~5}$ for CLIP. As for our method, we set $H$ in MAP to ${3,~4,~5}$ correspondingly and disable MCB. The results are shown in Tab.~\ref{map_vs_clip}. The learned motion-aware prompts exceeds hand-crafted prompts by over $+20\%$ Top-1 accuracy, showing the strong ability of our motion prompts learning in adapting CLIP to action recognition task. In other word, steered by motion cues, learning to understand the motion flowing in a video is much better than gazing at still frames.

\noindent\textbf{The initialization of MAP} We discuss the effects of the initialization for MAP. In the setting of $H=5$, we design $3$ hand-crafted prompts to initialize the learnable vectors in MAP. As shown in Tab.~\ref{init_map}, the performance is barely affected by the content of prompts for initialization, showing that the motion prompts learning is not only efficient, but also stable.

\noindent\textbf{Trainable parameters and GFLOPs} Our methods adapts CLIP into action recognition task with far few trainable parameters and additional computations. As shown in Tab.~\ref{param_flops}, our method merely leverages $4.2M$ trainable parameters and $0.03$ additional GFLOPs but achieves well adaptation of CLIP.

\begin{table}
\setlength{\tabcolsep}{1.2mm}{
    \begin{tabular}{ccccccc}
    % \hline
    \toprule
    \multirow{2}{*}{Methods} & \multicolumn{2}{c}{HMDB-51} & \multicolumn{2}{c}{UCF-101} & \multicolumn{2}{c}{Kinetics-400} \\ \cline{2-7} 
                  & Top-1 & Top-5 & Top-1 & Top-5 & Top-1 & Top-5 \\ \hline
    I3D~\cite{kay2017kinetics} & 74.3  & -     & 95.1  & - & 71.6  & 90.0  \\
    S3D-G~\cite{xie2018rethinking} & \textbf{75.9} & -     & \textbf{96.8} & -     & 74.7  & 93.4  \\
    R(2+1)D~\cite{tran2018closer} & 74.5  & -     & \textbf{96.8}  & -     & 72.0  & 90.0  \\
    TSM~\cite{lin2019tsm} & -     & -     & -     & -     & 74.7  & -     \\
    R3D-50~\cite{hara2018can} & 66.0  & -     & 92.0  & -     & -     & -     \\
    NL-I3D~\cite{wang2018non} & 66.0  & -     & -     & -     & 76.5  & 92.6  \\
    SlowFast~\cite{feichtenhofer2019slowfast} & -     & -     & -     & -     & 77.0  & 92.6  \\
    X3D-XXL~\cite{feichtenhofer2020x3d} & -     & -     & -     & -     & 80.4  & 94.6  \\
    TimeSformer-L~\cite{bertasius2021space} & -     & -     & -     & -     & \textbf{80.7}  & \textbf{94.7}  \\
    Ju~\textit{et al.}~\cite{ju2021prompting} & 66.4  & 92.1  & 93.6  & 99.0  & 76.6  & 93.3  \\ 
    \hline
     Ours &  72.9 &  93.2 &  96.3 &  99.3 &  77.4 &  93.6 \\
    \bottomrule
    \end{tabular}
}
\caption{Comparison to state-of-the-art methods on HMDB-51, UCF-101 and Kinetics-400 datasets. Our method achieves competitive performance leveraging extremely few trainable parameters and additional computational costs.}
\label{close_set}
\end{table}

\subsection{Few-Shot Training}
Here we conduct ``few-shot'' experiments on HMDB-51, UCF-101 and Kinetics-400 datasets. All models are trained on $8$ Tesla-V100 cards with a smaller batchsize of $32$ and an initial learning rate of $0.0015$ which can significantly stabilise the training procedure. We also discuss our three key elements, \textit{i.e.,} MMB, MAP, and MCB under ``few-shot'' training. For fair comparisons, we follow the standard $k$-shot protocol in X-CLIP~\cite{ni2022expanding} for HMDB-51 and UCF-101 datasets, where $k\in\{2,4,8,16\}$. We randomly select $k$ videos for every category to construct train set and evaluate the ``few-shot'' models on the standard test set. 
The results of Top-1 accuracy are reported as the average of $10$ runs. Meanwhile, we also compare our ``few-shot'' ability with state-of-the-art methods, including TSM~\cite{lin2019tsm}, TimeSformer~\cite{bertasius2021space}, Swin-B~\cite{liu2022video} and X-CLIP~\cite{ni2022expanding}. As shown in Tab.~\ref{few_shot_hmdb_ucf}, three conclusions are drawn as follows: \rmnum{1}) MMB, MAP and MCB still play key roles on ``few-shot'' training. Taking the $2$-shot model of HMDB-51 for instance, the model equipped with MCB yields a gain of $+13.7\%$ against ``Baseline'', while MMB improves $+17.4\%$. And the combination of MMB and MAP improves $+18.5\%$ on Top-1 accuracy. Finally, the integral model boosts the accuracy of ``Baseline'' by a margin of $+19.6\%$. \rmnum{2}) Our model gains more with a smaller $k$ compared with the state-of-the-art methods. For instance, leveraging around $1\%$ training data, the $2$-shot model of UCF-101 exceeds the best X-CLIP~\cite{ni2022expanding} by a significant margin of $+6.0\%$. However, the $16$-shot model of UCF-101 merely exceeds X-CLIP by $+0.2\%$. It needs to mention that, the $2$-shot models of HMDB-51 and UCF-101 with only MMB and MAP have already surpassed the best X-CLIP by $+1.2\%$ and $+2.0\%$ respectively, demonstrating the strong ``few-shot'' learning ability of our method with extremely limited training data, \textit{i.e.,} less than $2\%$ videos in the train set. \rmnum{3}) In summary, our method achieves new state-of-the-art ``few-shot'' performance on both HMDB-51 and UCF-101 datasets.

Following the setting of Ju~\textit{et al.}~\cite{ju2021prompting}, we also conduct a standard $5$-shot experiment on the Kinetics-400 dataset. We report the averaged Top-1 accuracy of $10$ runs in Tab.~\ref{few_shot_k400}. The ``Baseline'' model equipped with MCB has achieved a competitive Top-1 accuracy against Ju~\textit{et al.}. Furthermore, our integral model yields the best Top-1 accuracy and exceeds the state-of-the-art of Ju~\textit{et al.} by $+3.0\%$, using less than $1\%$ videos for training. These results show the excellent efficiency of our method.

\subsection{Zero-Shot Training}
Zero-shot training is challenging as the categories that need to predict are entirely unseen to the model. Following the experimental settings of \cite{ni2022expanding,chen2021elaborative,radford2021learning}, we adopt a two-stage strategy to verify the ``zero-shot'' ability of our method. Firstly, we train a model on Kinetics-400 dataset~\cite{kay2017kinetics} under the ``closed-set'' setting. The training is conducted on $32$ Tesla-V100 cards with a batchsize of $256$ and an initial learning rate of $0.0012$ for $10$ epochs. Secondly, we adapt the model to HMDB-51 and UCF-101 datasets without any additional tuning. Here, we report the average Top-1 accuracy and the standard deviation over three traditional splits of both HMDB-51 and UCF-101 in Tab.~\ref{zero_shot_k400}. We compare our ``zero-shot'' Top-1 accuracy with most existing ``zero-shot'' studies in action recognition. Our method exceeds the best state-of-the-art method, \textit{i.e.,} X-CLIP by $+5.5\%$ on HMDB-51 and $+4.4\%$ on UCF-101. Furthermore, we also conduct ``zero-shot'' training across HMDB-51 and UCF-101 datasets. The results are shown in Tab.~\ref{zero_shot_hmdb_ucf}. 
Compared to ``Baseline'', our model achieves a $+28.9\%$ gain on UCF-101 and a $+21.8\%$ improvement on HMDB-51. In summary, our model built upon CLIP still maintains robust ``zero-shot'' generalization to recognize unseen categories.

\subsection{Comparison to State-of-The-Art}
Finally, we compare our best models on ``closed-set'' training with state-of-the-art methods on three datasets. As shown in Tab.~\ref{close_set}, the Top-1 accuracy of ours exceeds the concurrent CLIP-like method Ju \textit{et al.} \cite{ju2021prompting} by $+6.5\%$, $+2.7\%$ and $+0.8\%$ on HMDB-51, UCF-101 and Kinetics-400 datasets, respectively. Our method also achieves comparable performance against CNN-based and ViT-based state-of-the-art methods leveraging extremely few additional training parameters ($4.2$~M) and computation costs ($0.03$~GFLOPs) and maintains strong generalization as mentioned before.

\section{Conclusions}
This paper focuses on enhancing the \textit{efficiency} and \textit{generalization} of action recognition. Based on Contrastive Language-Image Pre-training (CLIP), we discuss three critical problems, \textit{i.e.,} the modeling of motion information, the diversity of prompts, and the communication of multimodal representations. We first explicitly model the motion by the difference of frame-level representations. The captured motion information enhances the video representation and steers a dynamic prompts learner to generate more various prompts. Lastly, we utilize dual cross-modal attention to achieve collaborative learning. As a result, our proposed method shows a remarkable ``few-shot'' ability that exceeds most existing methods by a significant marge using extremely few training data on three datasets. Meanwhile, our approach also performs better in ``zero-shot'' transfer learning and yields a competitive performance against most of the state-of-the-art methods.

\bibliographystyle{ACM-Reference-Format}
\balance
\bibliography{sample-base}

\end{document}